\documentclass[a4paper,fleqn]{cas-dc}

\usepackage[authoryear]{natbib}
\usepackage{setspace}
\usepackage{tikz}
\usetikzlibrary{arrows.meta,positioning,fit,calc,backgrounds}
\usepackage{hyperref}

\usepackage{lineno}

\definecolor{tealblue}{rgb}{0.21, 0.46, 0.53}
\definecolor{tiffanyblue}{rgb}{0.04, 0.73, 0.71}
\definecolor{ballblue}{rgb}{0.13, 0.67, 0.8}
\definecolor{elsevierblue}{rgb}{0.21, 0.6, 0.92}
\hypersetup{citecolor=ballblue, linkcolor=Red, urlcolor=Blue}

\begin{document}
\let\WriteBookmarks\relax
\def\floatpagepagefraction{1}
\def\textpagefraction{.001}
\shorttitle{Super-resolution of Forest ALS Point Clouds}
\shortauthors{Shao et~al.}

\title [mode = title]{Super-resolution of airborne laser scanning point clouds for forest inventory}                      



\author[1]{Jinyuan Shao}

\author[2]{Sangyoong Park}

\author[2]{Chunxi Zhao}

\author[2]{Ayman Habib}

\author[1]{Songlin Fei}[orcid=0000-0003-2772-0166]
\cormark[1]
\ead{sfei@purdue.edu}


\affiliation[1]{organization={Department of Forestry and Natural Resources, Purdue University},
                country={USA}}
\affiliation[2]{organization={Lyles School of Civil and Construction Engineering, Purdue University},
                country={USA}}








\cortext[cor1]{Corresponding author}


\begin{abstract}
Airborne Laser Scanning (ALS) can collect point clouds across large areas, enabling large-scale forest inventory. However, ALS point clouds are sparse and noisy, resulting in inaccurate individual-tree-level forest inventory, such as stem localization and tree size estimation. To overcome this problem, we propose a deep learning model, 3D Forest Super Resolution (3DFSR),  to simultaneously improve point density and reduce noise for ALS forest point cloud. 3DFSR is a voxel-based Convolutional Neural Network (CNN) with a U-Net architecture. The proposed 3DFSR is evaluated on ALS point clouds collected in both temperate forests in the U.S. and boreal forests in Germany. Terrestrial Laser Scanning (TLS) and Mobile Laser Scanning (MLS) point clouds are used as references to evaluate the performance. Experimental results demonstrate that 3DFSR can generate finer point clouds of tree structure than other state-of-the-art point cloud super-resolution algorithms, achieving 0.249 m Chamfer Distance and 2.711 m Hausdorff Distance. Furthermore, to verify the effectiveness of super-resolution point clouds in forest inventory, we conduct experiments of stem detection, diameter at breast height (DBH) measurements, and stem reconstruction on both original ALS point clouds and 3DFSR enhanced point clouds. We find that stem detection and reconstruction algorithms developed for TLS/MLS point clouds can directly work on our 3DFSR point clouds, and DBH can be derived with circle-fitting method from the 3DFSR point cloud. F1 score of stem detection is improved from 0.71 on original ALS point clouds to 0.97 on super-resolution point clouds; DBH estimation improves from 13.45 cm Root Mean Square Error (RMSE) using allometric equations to 6.43 cm using circle fitting; comparing to stems reconstruction from MLS point clouds, stem reconstructed from 3DFSR point clouds has 0.170 m of Chamfer Distance and 0.377 m of Hausdorff Distance, and 0.95 R\textsuperscript{2} volume estimation. Finally, we find that the proposed 3DFSR is applicable to process point densities from 10 to 1700 points/m\textsuperscript{2}; it also can be generalized across data collected from different LiDAR platforms without transfer learning. We hope that this study can expand the application of ALS point clouds in forest analysis.
\end{abstract}




\maketitle

\section{Introduction}

Forests are essential to human well‑being by regulating climate, storing and cycling carbon, sustaining hydrological and nutrient flows, and providing timber, food and cultural benefits. Yet robust estimation of  large-scale ecological and economic values they deliver remains uncertainty because most assessments still rely on coarse plot‑scale inventories. In addition, rapid global changes are reshaping forest composition, productivity and disturbance regimes, and thus impacting the services and revenues that forests supply. Therefore, to accurately analyze and estimate forest ecological and economic values and to monitor their changes, it is necessary to obtain timely and spatially explicit individual tree-level forest inventory data.

Airborne Laser Scanning (ALS) technique provides an opportunity for large-scale forest inventory. It captures georeferenced 3D point clouds to represent forests, enabling detailed and nondestructive forest measurements  across a large geographic area. ALS has been widely explored for various forest inventory tasks, such as crown delineation, height measurements, species recognition, and biomass estimation~\citep{dalponte2014tree, wang2019field, wang2024individual, oehmcke2024deep}. These applications demonstrate the capability of ALS for forest inventory, and to provide spatially continuous and repeatable measurements across vast regions.

However, the low density of ALS point clouds limits further detailed individual-tree-level forest inventory. ALS has a long sensing range to target objects and the reflected laser signal decreases with distance, making it more susceptible to noise and potentially leading to inaccurate or missed measurements~\citep{soudarissanane2011scanning}. Although some studies have explored individual-tree-level forest inventory using ALS point clouds (e.g., ~\cite{parkan2015individual} used ALS point clouds with 50 points/m\textsuperscript{2} for tree segmentation, and~\cite{shao2023robust} introduced a clustering approach for tree detection on 3DEP (3D Elevation Program) data), ALS data often remain sparse and noisy. Such point distributions frequently result in incomplete representation of trees, especially stems, making it difficult to extract fine-scale structural attributes such as accurate stem locations and diameter at breast height (DBH). This limitation restricts the performance of downstream tasks, including stem mapping and DBH estimation. 

Enhancing the density and quality of ALS forest point cloud can improve its applicability. Most studies on individual-tree-level inventory focus on the use of close-range sensing platforms such as Terrestrial Laser Scanning (TLS) and Mobile Laser Scanning (MLS)~\citep{liang2016terrestrial, shao2024large}. This is because they have relatively short range of sensing to the target objects, so the captured point cloud has high point density and the shape of the tree can be captured well. Research has shown TLS/MLS can achieve more accurate forest inventory results compared to ALS~\citep{kankare2014accuracy, lin2022comparative}. And the major reason is that a denser and more complete point clouds would provide more information and enable reliable detection of stems and more accurate DBH estimation. Therefore, improving the density of ALS point clouds is a critical step toward scalable and fine-resolution forest inventory at regional to continental scales.

There is no super-resolution algorithm developed for ALS point clouds for forest environment. Existing point cloud upsampling or super-resolution methods are primarily designed for synthetic objects or vehicle-based urban scenes scanning~\citep{zhang2022point, nunes2024scaling}, where geometric structures are relatively regular and well-defined. In contrast, forest environments exhibit irregular and unorganized characteristics, with complex interactions among stems, branches, and understory, making them different from synthetic objects or urban scenes. In addition, ALS point clouds often have uneven point densities across different parts of a tree, a high level of noise, and sometimes incomplete scannings. Directly applying generic point cloud processing models to ALS forest data may lead predicted point clouds with uneven point distribution with added noise and cannot produce complete tree shapes. Some studies have explored individual tree point cloud completion~\citep{xu2023single, bornand2024completing, luo2026inceptionformer}; however, they focus specifically on the completion of missing parts and are not applicable to scene-level upsampling and denoising. Consequently, a forest-specific super-resolution algorithm is needed that accounts for challenging forested environments and low-quality ALS point clouds, recovers stem geometry, and enhances structural details while maintaining consistency with the original ALS observations.

In this study, we propose 3D Forest Super Resolution (3DFSR), a super-resolution algorithm for ALS forest point clouds to achieve more accurate forest inventory. The core concept of this model is to convert coordinate regression problem into voxel occupancy problem. The core module of our algorithm is a sparse convolutional U-Net that generates dense and clean point clouds from sparse and noisy ALS point clouds. Our algorithm can produce super-resolution point clouds whose density is comparable to TLS data and improve the quality and applicability of ALS point clouds. A comprehensive comparison between our algorithm and other representative approaches is conducted on ALS point cloud collected in three plots in temperate forests in the U.S. and two plots in boreal forests in Germany. In addition, we extensively compare the performance before and after point cloud super-resolution in stem detection and DBH estimation; and verify its potential on stem reconstruction. Finally, we validate the model's capability regarding input point density by downsampling high-resolution UAV LiDAR data to different point densities; and explore its generalization across LiDAR platforms. To the best of our knowledge, this is the first study to explore super-resolution for scene-level ALS forest point clouds.
\section{Related work}

In this section, we review the research on retrieving forest attributes using ALS point clouds. Then, we introduce the related work on point cloud super-resolution.

\subsection{ALS point clouds for forest inventory}

Research using ALS for individual-tree-level forest inventory can be categorized as implicit predictions and explicit measurements.

{\bf Implicit predictions.} Implicit predictions take a point cloud or its attributes as input, and build a regression relationship to predict biometrics, such as number of trees, height, DBH, stem volume and biomass.~\cite{salas2010modelling, jucker2017allometric} developed allometrtic models using tree height and crown diameter from ALS point clouds to estimate individual tree's diameter or biomass.~\cite{apostol2020species} developed a linear regression model to predict DBH with ALS derived height and crown diameter.~\cite{yu2011predicting} used a Random Forest to predict single tree's height, DBH, and stem volume with attributes extracted from tree point clouds.~\cite{briechle2020classification, wang2024individual} developed deep learning models to classify tree species.~\cite{oehmcke2022deep,oehmcke2024deep} used deep learning methods to predict biomass, wood volume, and carbon stocks for the entire forest from ALS point cloud. To achieve implicit predictions, annotated data are needed to train the model. In addition, for forest-level inventory, such as stem mapping, implicit predictions cannot provide explicit locational information (e.g., tree locations)

{\bf Explicit measurements.} Explicit measurements directly detect or fit biometrics from point clouds. There have been a lot of studies showing tree height can be directly measured from ALS point clouds~\citep{erfanifard2018development, wang2019field, lin2022comparative}, as laser pulses are not occluded before they encounter the tree top, so there is a relatively accurate height value~\citep{wang2019field}. For number of trees, explicit methods conduct individual tree detection based on crown segmentation. For example,~\cite{li2012new} develop an algorithm for individual tree segmentation based on geometry relationship between trees.~\cite{duncanson2014efficient, apostol2020species} utilize CHM to detect individual trees. The problem of crown-based methods is that they cannot deal with suppressed trees under dominant or co-dominant trees~\citep{cao2023benchmarking}. Detecting stems can solve this problem, ~\cite{hyyppa2022direct} develop a direct measurement method to detect arcs to conduct stem detection.~\cite{amiri2017detection} develop a shape descriptors based method to detect single tree stems. However, due  the sparsity of ALS point clouds, more detailed attributes, such as DBH, cannot be directly measured from ALS point clouds like TLS point clouds.

In summary, both implicit predictions and explicit measurements have been used for individual-tree-level forest inventory, but implicit predictions often are supervised and need additional reference data to fit models, while explicit measurements are unsupervised without model fitting so attributes can be directly derived from point cloud itself. However, studies have shown that sparse ALS point clouds result in less accurate of tree inventory~\citep{kankare2014accuracy, lin2022comparative}. Therefore, increasing ALS point cloud density can potentially improve forest inventory performance.

\subsection{Point cloud super-resolution}

Point cloud super-resolution, also known as point cloud upsampling, is one of the fundamental point cloud analysis tasks. It takes the low-resolution point cloud as input and generates a high-resolution point cloud with details. Existing methods can be divided into geometry-based and learning-based.

{\bf Geometry-based.} Early works resample points from surfaces reconstructed from sparse points to achieve upsampling. For example,~\cite{alexa2003computing} interpolates points at vertices of Voronoi diagram in the local tangent space.~\cite{lipman2007parameterization, huang2009consolidation} present and adapt locally optimal projection (LOP) operator for surface reconstruction so that point cloud can be upsampled. Further,~\cite{huang2013edge} propose an edge-aware method to resample points from edges to corners, with the help of normal (a vector perpendicular to the surface). Geometry-based methods often unsupervised, meaning it does not need additonal data to train models. However, geometry-based methods heavily relies on the accuracy of the normals at given points and geometric priors.

{\bf Learning-based.} Learning-based point cloud upsampling methods have been widely explored in recent years.~\cite{yu2018pu} first developed PU-Net for point cloud upsampling task. PU-Net extracts multi-scale features for each point and expands them to achieve the target upsampling ratio, and then reconstructs point coordinates from feature layers. Following this pattern of feature expansion, subsequent methods have achieved better results~\citep{li2019pu, yifan2019patch, li2021point, rong2024repkpu}.~\cite{yifan2019patch} developed a network to progressively expand feature layers and the point cloud is upsampled gradually;~\cite{li2021point} proposed a spatial refiner after feature expansion for detailed coordinate reconstruction. Another kind of method first conducts unsupervised interpolation on a sparse point cloud, and then refines the coordinates for the interpolated point cloud. The pioneering work is Grad-PU, which first proposes a midpoint interpolation method and then develops a distance regressor for coordinate optimization~\citep{he2023grad}. Subsequent studies focus on improving coordinate optimization. For example, PUDM is a conditional denoising diffusion probabilistic model for point cloud upsampling based on interpolated points~\citep{qu2024conditional};~\cite{liu2025efficient} proposed a flow matching approach to upsample point clouds. One of the advantages of this  method is that it can achieve arbitrary number of points for upsampling, avoiding the upsampling module design~\citep{he2023grad}.

Overall, geometry-based methods are unsupervised but heavily rely on prior knowledge; learning-based methods have shown great potential, but their performance is unknown on ALS point clouds with uneven density and high noise, especially in forest environments with irregular shape of trees and unordered structures, which limits the potential of using ALS point clouds to conduct detailed forest inventory.
\section{Materials}

We describe the data used in this section, including study sites, data collection and attributes, dataset creation, and reference data.

\subsection{Study sites}
\label{study_sites}
\begin{table*}[t]
\tiny
\label{tab:plot}
\centering
\caption{Study sites and their information.}
\resizebox{\textwidth}{!}{%
\begin{tabular}{@{}lcccccc@{}}
\toprule
         & Country & Forest type       & Dominant species & \# of stems & Stem density (stems/ha) & Avg. DBH (cm) \\ \midrule
Plot I   & U.S.    & Plantation        & Black walnut     & 91          & 116      & 39.10       \\
Plot II  & U.S.    & Mixed deciduous   & Oak; beech       & 30          & 480       & 37.09       \\
Plot III & U.S.    & Mixed deciduous   & Oak; maple       & 80          & 320      & 34.50       \\
Plot IV  & Germany & Coniferous        & Douglas fir      &  9          & 44       & 69.21       \\
Plot V   & Germany & Mixed             & Pine; oak        & 11          & 52      & 54.60       \\
\bottomrule
\end{tabular}%
}
\end{table*}

Five plots are selected to conduct our research, three of which are in the U.S. and two in Germany. Plot I, II and III are located in West Lafayette, Indiana, USA, and Plot IV and V are located in Baden-Wuerttemberg, Germany.

{\bf Plot I.} Plot I (40°25'54"N 87°02'24"W) is a black walnut (\textit{Juglans nigra}) plantation established in 2007, which is located in West Lafayette, Indiana, USA.There are 92 trees in this plantation with a maximum height of 26.3 m and an average DBH of 39.10 cm. It has regular planting pattern on a 5 X 5 m spacing and there is no understory vegetation, so it is a forest with a simple structure.

{\bf Plot II.} Plot II (40°25'43"N 86°56'15"W) is a forest research park located on the edge of the Purdue University campus. It has area of 24 ha of natural forest with mixed deciduous overstory trees and some understory vegetation. We selected a plot with 20 m radius as one of our study sites. In this plot, there are 16 trees with oak (\textit{Quercus} spp.) and beech (\textit{Fagus grandfolia}) as the main species. The maximum tree height is 35 m and the average DBH is 37.06 cm.

{\bf Plot III.} Plot III (40°26'14"N 87°02'12"W) is located at Martell Forest in West Lafayette, IN, US. This is a natural forest with an area of 8 ha. It has at least 20 overstory tree species. We select an area with 70 by 150 m. In this plot, there are overall 75 trees with an average DBH of 34.5 cm, and a maximum height of about 40 m. This is a temperate natural forest with a dense understory and complex structure. 

{\bf Plot IV.} Plot IV (49°01'01"N 8°41'12"E) is located in Bretten, Baden-Wuerttemberg, Germany. In this area, we select a plot of 45 m radius. It is a coniferous forest and contains only Douglas fir (\textit{Pseudotsuga menziesii} ). The maximum tree height is around 52 m and the average DBH is 69.21 cm. The terrain has an 8 degree slope. More information can be found in~\cite{weiser2022individual}

{\bf Plot V.} Plot V (49°01'47"N 8°25'08"E) is located in Karlsruhe, Baden-Wuerttemberg, Germany. We also select square plot with 45 by 45 m. This is a mixed forest with both conifers and deciduous trees, with pine (\textit{Pinus} spp.) and oak as the main species. The maximum tree height is around 30 m and the average DBH is 54.6 cm. Detailed in formation can be found in~\cite{weiser2022individual}.

\subsection{Data collection}
\label{data_collection}

\begin{figure}[]
	\centering
	\includegraphics[width=.99\columnwidth]{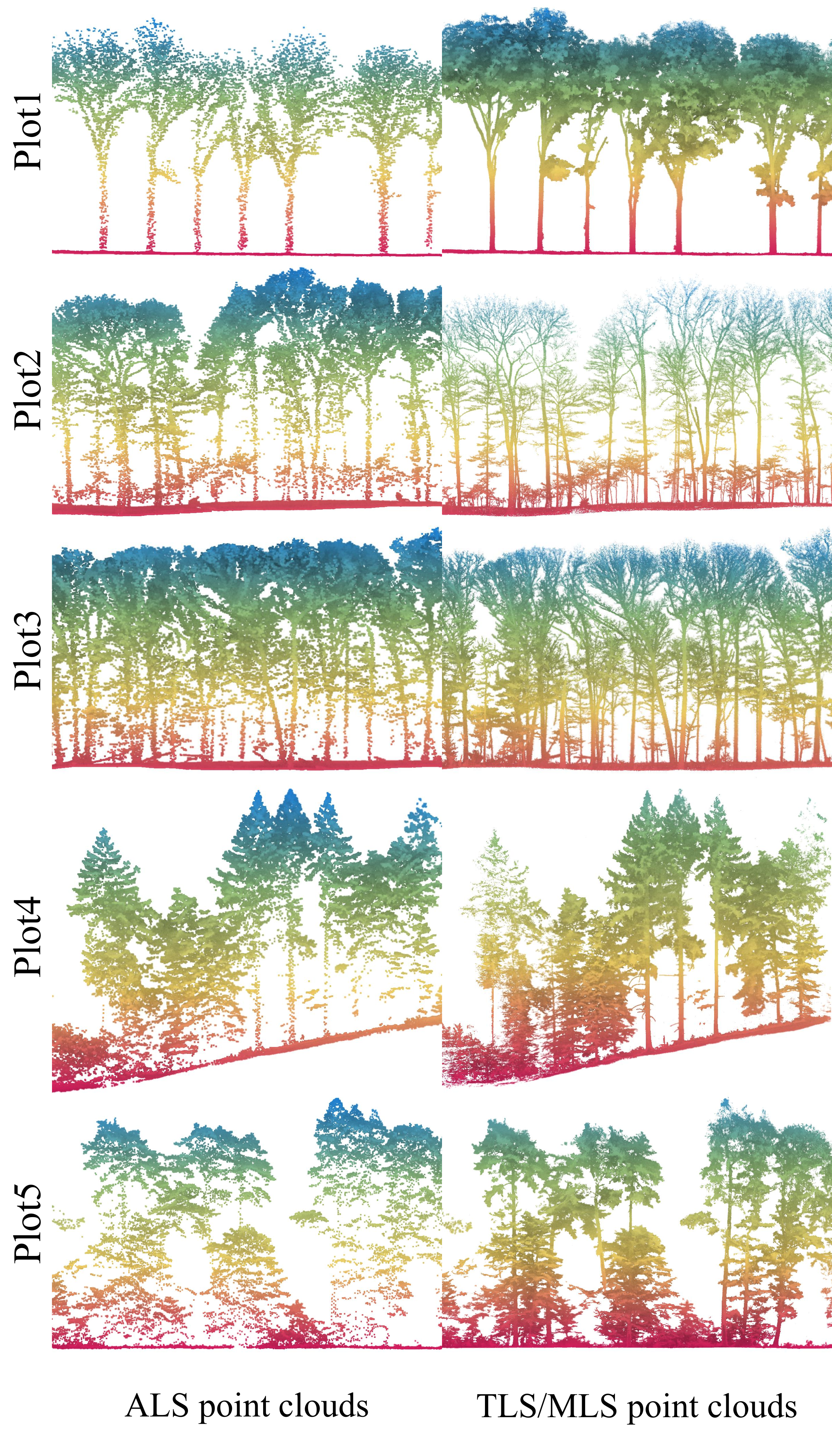}
	\caption{Examples of ALS and TLS/MLS LiDAR point clouds used in this study.}
	\label{fig: data}
\end{figure}

We introduce data collection of low-resolution ALS data and high-resolution TLS/MLS data used in this study. Fig.~\ref{fig: data} shows examples of ALS and TLS/MLS point clouds data for each plot, and Tab.~\ref{tab2} presents their attributes.  
\begin{table*}[]
\centering
\caption{Data characteristics of low-resolution of ALS and high-resolution TLS/MLS data.}
\label{tab2}
\resizebox{\textwidth}{!}{%
\begin{tabular}{@{}lcccccccc@{}}
\toprule
       & \multicolumn{4}{c}{ALS} & \multicolumn{4}{c}{TLS/MLS} \\ \midrule
 & Sensor type & Flying height (m) & Data acquisition time & Avg. point density (pts/m\textsuperscript{2}) & Sensor type & Range (m) & Data acquisition time & Avg. point density (pts/m\textsuperscript{2}) \\ \midrule
Plot I   & Geiger-mode  & 3700  & 12/12/2021  & 102 & MLS     & 100  & 9/5/2024      & 7122    \\
Plot II  & Geiger-mode  & 3700  & 12/12/2021  & 185 & MLS     & 100  & 3/2/2023      & 6032    \\
Plot III & Geiger-mode  & 3700  & 12/12/2021  & 190 & MLS     & 100  & 3/4/2024      & 8451    \\
Plot IV  & Liner-mode   & 140   & 7/5/2019    & 134 & TLS     & 600  & 7/4/2019      & 46321   \\
Plot V   & Liner-mode   & 300   & 7/5/2019    & 148 & TLS     & 600  & 7/30/2019     & 46478   \\ \bottomrule
\end{tabular}%
}
\end{table*}

\subsubsection{ALS data collection.}

ALS data for Plot I, II and III were collected by the ALS system developed by VeriDaas Corporation (Denver, CO, USA). This system contains a Geiger-mode LiDAR sensor and an Applanix POS AV 610 for direct georeferencing. The Geiger-mode LiDAR sensor consists of arrays of Geiger-mode Avalanche Photodiode (GmAPD) detectors. Each of the GmAPD detectors is capable of detecting the return signal with a few photons~\citet{clifton2015medium}. The extreme sensitivity of GmAPD detectors allows the design of LiDAR systems that operate at a lower energy, higher altitude, and faster flying speed, and acquire measurements in a much higher density compared to linear LiDAR systems. The VeriDaaS system has an array of 32 by 128 GmAPD detectors, which effectively collects 204,800,000 observations per second with a pulse repetition rate of 50 kHz. The use of a Palmer scanner, together with a 15-degree scan angle of the laser and scan pattern of a 50\% swath overlap, enables multi-view data collection. The data were collected in leaf-off season December 2021 at a height of around 3700 m elevation. The final data product has an overall point density of 150 points/m\textsuperscript{2}.

ALS data for Plot IV and V were acquired using a RIEGL VQ-780i full-waveform laser scanning system mounted on a Cessna C207 aircraft. The flight missions were conducted on 5 July 2019 under leaf-on conditions, consisting of parallel flight strips supplemented by two orthogonal cross strips to ensure uniform coverage. The point clouds were georeferenced to ETRS89 / UTM zone 32 (EPSG:25832) with ellipsoidal heights (GRS80). More details can be found in~\cite{weiser2022individual}.

\subsubsection{TLS/MLS data collection.}

The high-resolution data for Plot I, II, and III were collected using an MLS backpack system, Hovermap ST, developed by Emesent. It is equipped with a 16 channels Velodyne LiDAR sensor with range from 0.4-100 m. To obtain a high-quality point cloud without occlusion, we walked through the forest with a zigzag trajectory and turned around at the edge of the forest to cover the target area when collecting data. The final data product was not georeferenced, so we manually registered it with the ALS data described above. The detailed characteristics of the data are shown in Tab~\ref{tab2}.

High-resolution data for Plot IV and V were collected using a RIEGL VZ-400 terrestrial laser scanner (1550 nm wavelength, 0.3 mrad beam divergence) operated at 300 kHz PRF with 0.017° angular resolution. The scanner was deployed at 5–8 scan positions per survey, regularly distributed around selected target trees within target forest plots, with occasional tilted scans to capture full tree height; cylindrical and circular reflectors were installed for multi-scan registration. Initial georeferencing was performed using RTK-GNSS measurements from one main scan position and one tie point, followed by tie-point registration and ICP-based refinement in RiSCAN PRO. Because GNSS accuracy under canopy was limited, the TLS point clouds were further georeferenced by registering them to ULS point clouds via tree matching and ICP alignment, including height-offset correction, to achieve accurate spatial positioning. Data collection were conducted between June and September 2019 under leaf-on conditions. More details can be found in~\cite{weiser2022individual}.

\subsection{Datasets for super-resolution}
\label{sec:sr datasets}
A high quality dataset is the prerequisite to train high performance deep learning models. In this study, the mode input is defined as low-resolution point cloud, and the output is defined as super-resolution point cloud which stems from high-resolution TLS/MLS point cloud. Therefore, low-resolution and high-resolution point clouds need to be spatially matched. Although they are already georeferenced during the data collection, to make sure the dataset quality, we manually conducted ICP using CloudCompare to make sure these two kinds of point clouds are spatially matched. In addition, due to differences between ALS and TLS/MLS data collection, TLS/MLS data often have very dense understory points, while ALS data don't due to the occlusion of the canopy. Our major goal is to recover existing ALS points, so the understory points collected from TLS/MLS need be excluded. Therefore, we use the method proposed by~\cite{shao2026three} to remove understory points, because we want to make model learn how to improve existing ALS points, instead of imaging the structure of the understory vegetation. It should be noted that none of the testing data are used during the model training. All data described in~\ref{data_collection} are for testing, and training data are from other areas described in~\cite{shao2024large} and~\cite{weiser2022individual}.

\subsection{Field measurements}
\label{sec:fi datasets}

For Plot I, II and III, tree locations are recorded by RTK in the field. Then they are imported to CloudCompare and visually check the locations with georeferenced MLS point cloud data. DBH is measured using a diameter tape at 1.37 m (4.5 ft) above ground. 

For Plot IV, and V, tree locations are defined as the intersection of the stem centerline with the ground surface and derived by slicing the lowest part of the TLS point clouds to estimate the stem center in the x–y plane, and results are visually checked and corrected when necessary. DBH is measured in the field at 1.3 m above ground using a DBH tape. Crown diameter is estimated from the dripline of the crown, measured in two orthogonal directions using a Haglöf Vertex-IV hypsometer, and averaged to obtain the final crown diameter.
\section{Method}
\label{method}

A deep learning model is developed to generate super-resolution point clouds. In this section, we introduce the model's architecture and training details.

\subsection{3DFSR}

\begin{figure*}[]
	\centering
	\includegraphics[width=.99\textwidth]{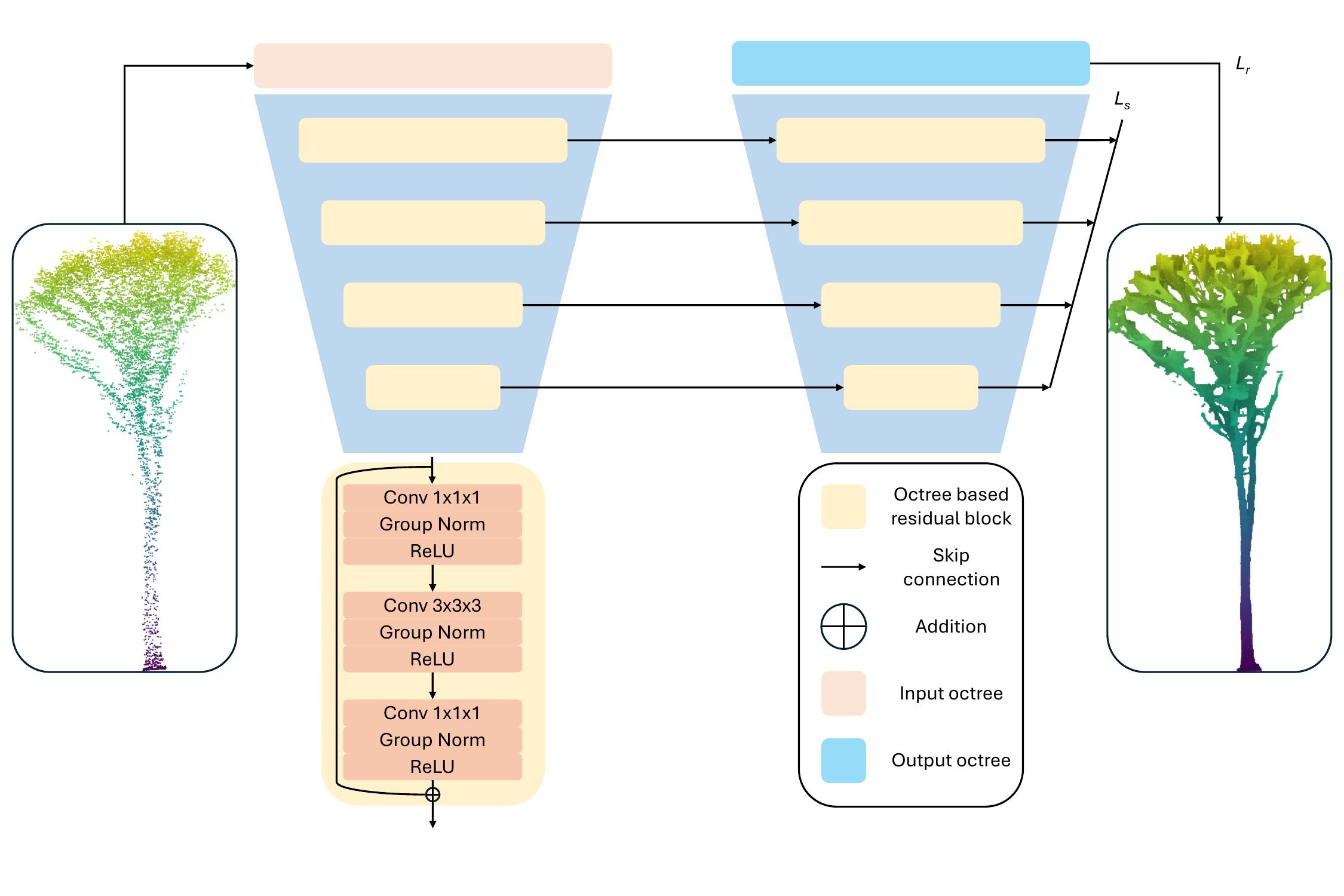}
	\caption{Model architecture of 3DFSR.}
	\label{fig: model}
\end{figure*}

We develop an octree-based UNet model to predict super-resolution point clouds. This method converts point cloud into octree, and learns features via octree-based CNN, then generates points on octree. Unlike other techniques that upsample and denoise separately, it simultaneity conducts denoising and upsampling. Fig.~\ref{fig: model} shows details of 3DFSR model architecture.

\begin{figure*}[]
	\centering
	\includegraphics[width=.99\textwidth]{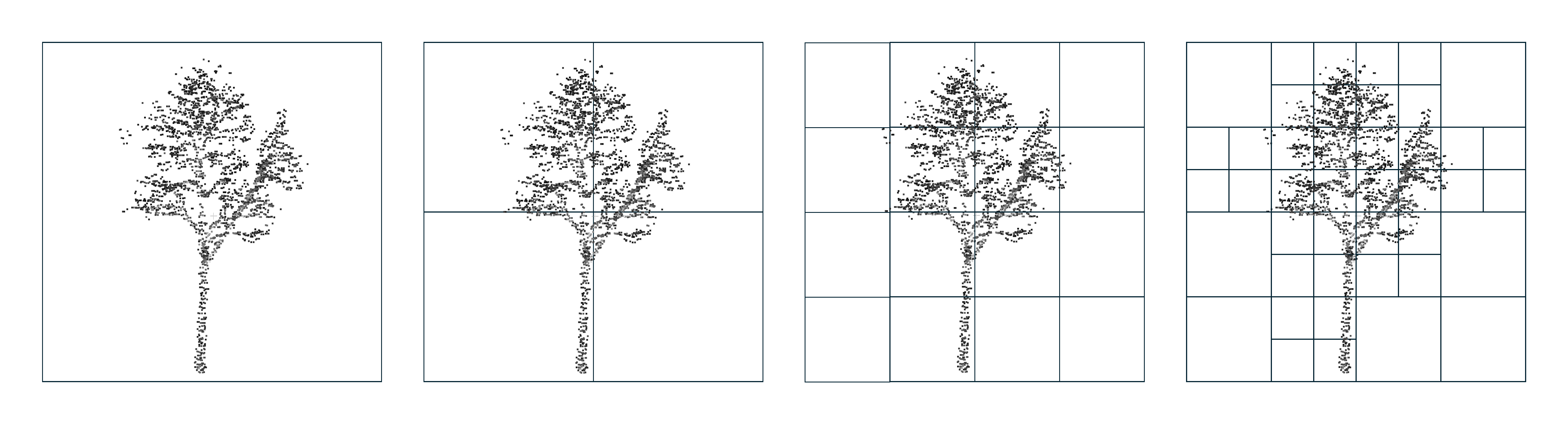}
	\caption{Octree-based voxelization.}
	\label{fig: octree}
\end{figure*}

{\bf Octree construction.} We first construct an octree on the input point cloud. Octree is a hierarchical tree data structure used to partition 3D space by recursively subdividing a volume into eight smaller child nodes. Each tree has a root node and eight child nodes, and each child node can serve as the root node of the next level of the octree~\citep{meagher1982geometric}. Therefore, the core of building an octree is to determine the root node and the child nodes. Fig.~\ref{fig: octree} shows an illustration of an octree construction for a forest point cloud. We first calculate the bounding box of the input point cloud, which serves as the root node of the octree; then, we divide the point cloud that belongs to the root node into eight cubes of equal size, called octants. Each octant is considered as a child node that belongs to the current root node. Then, for each non-empty octant, we conduct the same discipline of construction, i.e., divide the octant into eight child octants again, and regard each octant as a child node. The depth of the octree increases by 1 with each octree division. To achieve a balance between computational efficiency and point density, we use a depth of 8 in this study. 

{\bf Encoder.} The encoder follows a bottom-up hierarchical feature extraction process. At the coarsest resolution (depth $D$), a $1 \times 1 \times 1$ octree convolution projects the raw input features into a higher-dimensional latent space. Each level then applies two residual residual blocks, where each residual block consists of three consecutive octree convolutions with group normalization and ReLU activation, plus a skip connection that adds the block input to its output. This design helps stabilize training and preserves fine-grained geometry features. Between successive depths, we use downsampling blocks implemented as octree convolutions with kernel size $2$ and stride $2$. These blocks halve spatial resolution while increasing the dimensionality of the features, effectively aggregating the context of a larger receptive field. The resulting encoder representation captures both local shape details and global structural patterns.

{\bf Decoder.} The decoder reconstructs high-resolution representations through a symmetric top-down pathway. For each depth level, we first use an upsampling block implemented as an octree transposed convolution with kernel size $2$ and stride $2$ to double the spatial resolution. The upsampled features are then fused with encoder features from the corresponding depth of octree. This preserves high-frequency details that may otherwise be lost during downsampling. The fused features are refined by another residual block to improve spatial consistency before producing predictions. It should be noted that the skip connection happens on nodes are non-empty on both encoder and decoder. At each octree level, a shared prediction module decides whether a node is empty or occupied. Nodes identified as non-empty are subdivided into eight octants, and their features are passed down to the child nodes. This process is repeated until the target octree depth is reached. At the finest level, displacements are predicted for each non-empty node to represent the final geometry.

{\bf Loss.} In terms of the loss function, we combined structure loss and regression loss, denoted as Ls and Lr. The structure loss is to optimize the existing nodes of the octree, while the regression loss is to optimize the overall shape of the point cloud.

\begin{equation} \label{eq_loss}
L=\sum_{l=2}^d L_s^l+L_r,
\end{equation}

\subsection{Model training}

Training large point cloud scenes requires a significant amount of memory, which can exceed the capacity of GPUs. Therefore, dividing the entire point cloud scene into smaller blocks makes the training process computationally feasible. We split the entire point cloud scene into blocks with a spatial size of 10 m \texttimes 10 m \texttimes 10 m. Each of the blocks can be seen as an input data sample for deep learning models. For each block, we normalize its coordinates to [-1, 1]\textsuperscript{3} while keeping its shape and relative distances between the points. Specifically, we first move the block to the origin (0,0,0) by subtracting its centroid, then we divide all points by half of its bounding box size (5 m in this case). Point cloud coordinates normalization is a common preprocessing step for neural networks~\citep{shao2026three}. Since the learnable network parameters are randomly initialized in such a way that the distribution of the input data is preserved in the output, that preprocessing leads to faster convergence during optimization.

During model training, we use the AdamW optimizer to update the parameters with initial learning rate of 0.0005 and weight decay of 0.05. In addition, we use lambda learning rate to adjust the learning rate during training. We train our model with 2000 epochs, and use early stop technique to choose the best model (model that has best performance on validation set). 
\section{Experiments and Results}

In this section, we introduce six experiments to verify model's performance from different perspective, including super-resolution quality, stem detection, DBH estimation, stem reconstruction, effects on point density variation, and model transferability. For each experiment, we first introduce baselines and evaluation metrics being used, and then present experimental results.

\subsection{Super-resolution quality}
\label{sec:sr_quality}
\subsubsection{Experimental setting}
We use a set of methods and metrics to conduct this experiment to evaluate the quality of super-resolution, For baselines, we choose bilinear interpolation, PU-Net, GradPU. Bilinear interpolation is a kind of unsupervised method for point cloud super-resolution, PU-Net is the pioneer work to do point cloud upsampling with deep learning using feature expansion, and GradPU is representative work that follows paradigm of interpolation-refinement. All algorithms mentioned above and the proposed 3DFSR are run on the five plots mentioned in Section~\ref{study_sites}. It should be noted that PU-Net and GradPU can only achieve upsampling by a fixed rate (\textit{i.e.}, four times in this experimental setting), while 3DFSR produces output with a fixed voxel size, without requiring the upsampling rate to be set. Therefore, to ensure a fair comparison, we downsample the output point cloud of 3DFSR to four times the number of input points. In terms of metrics, we use the Chamfer distance and Hausdorff distance. These metrics are commonly used to evaluate the similarity between two point clouds, where lower values indicate better reconstruction quality.

\subsubsection{Super-resolution results}
\begin{figure*}[]
	\centering
	\includegraphics[width=.99\textwidth]{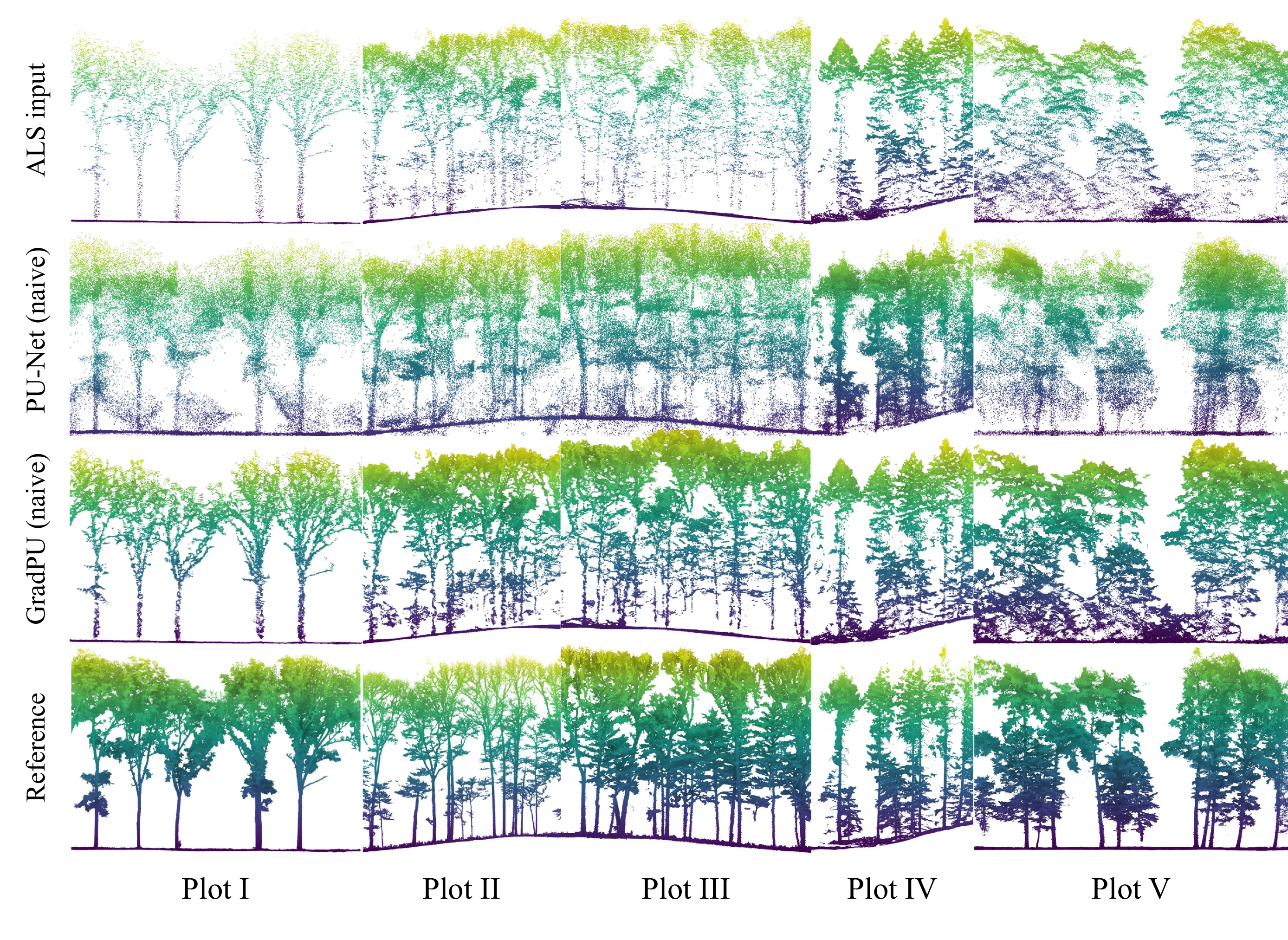}
	\caption{Super-resolution results obtained from naive models of PU-Net and GradPU in five plots.}
	\label{fig:sr_naive_plots}
\end{figure*}
\begin{figure*}[]
	\centering
	\includegraphics[width=.99\textwidth]{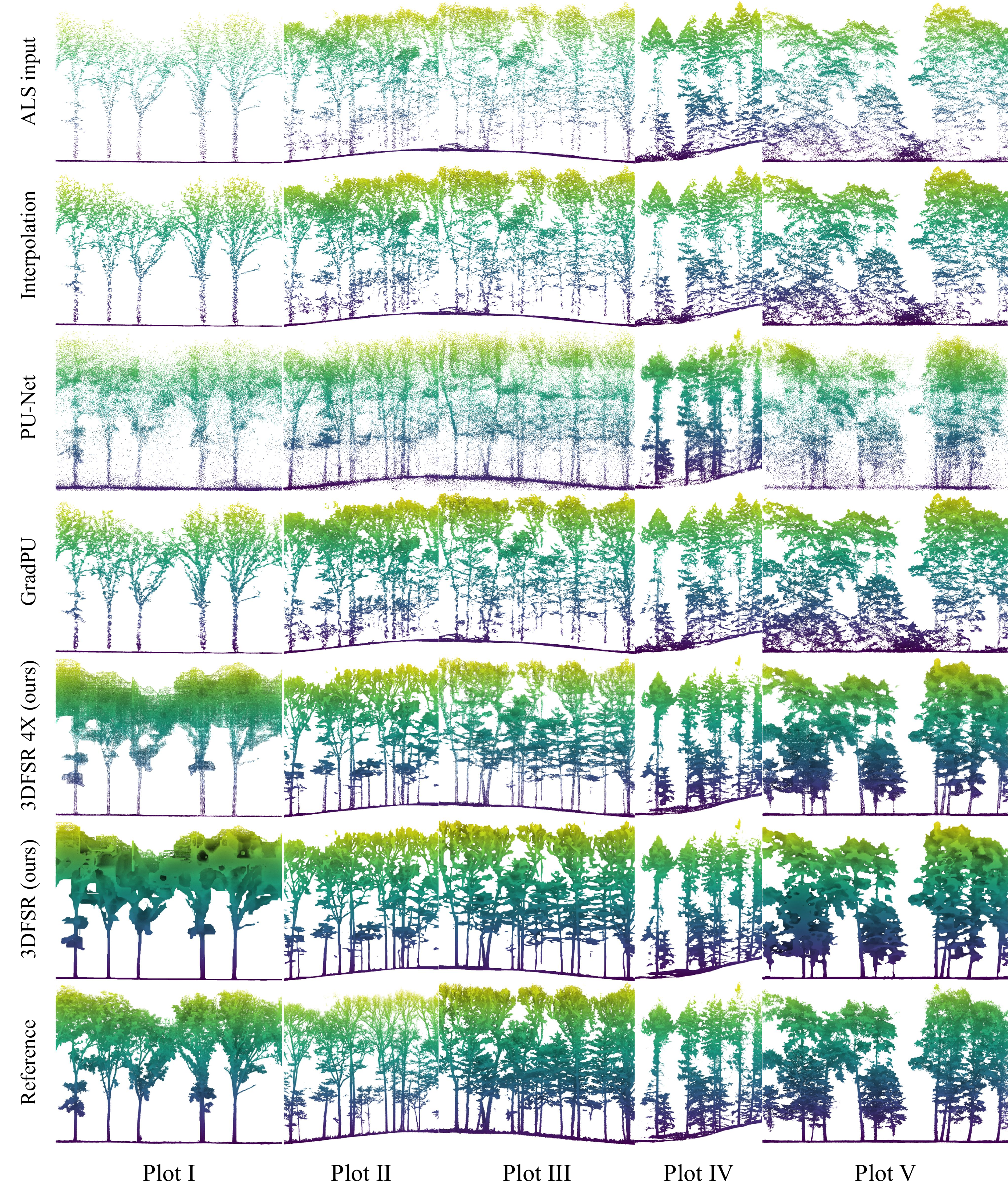}
	\caption{Super-resolution results with different methods in five plots.}
	\label{fig:sr_plots}
\end{figure*}
\begin{figure*}[]
	\centering
	\includegraphics[width=.99\textwidth]{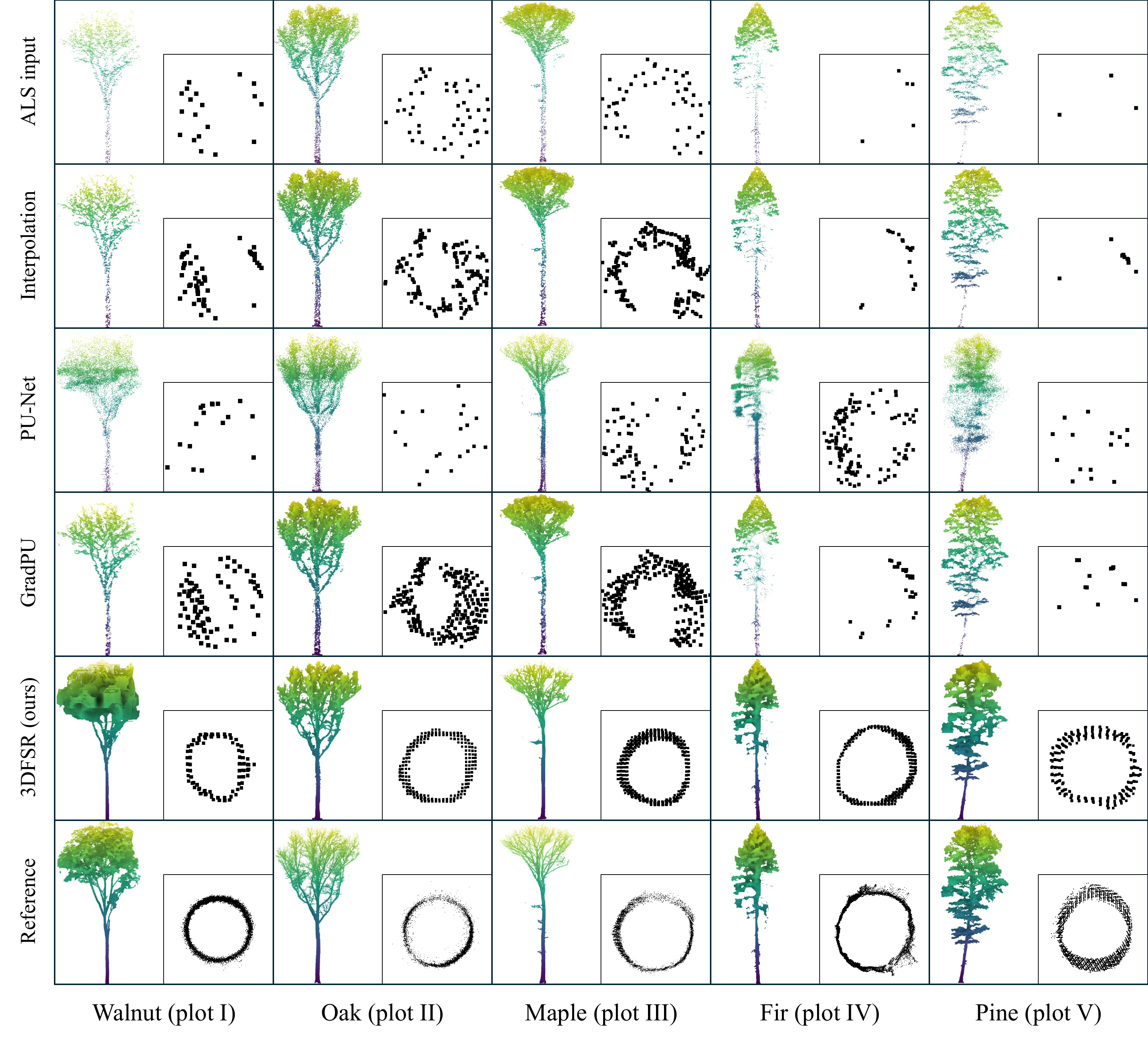}
	\caption{Super-resoltuion results with different methods on single trees of dominant species in five plots.}
	\label{fig:sr_singletree}
\end{figure*}

\begin{table*}[t!]
\tiny
\centering
\caption{Super-resolution quality comparison between different methods. The proposed 3DFSR achieves best Chamfer Distance (CD) and Hausdorff Distance (HD).}
\label{tab:sr_res}
\resizebox{\textwidth}{!}{
\begin{tabular}{@{}lcccccccccc@{}}
\toprule
              & \multicolumn{2}{c}{Plot I} & \multicolumn{2}{c}{Plot II} & \multicolumn{2}{c}{Plot III} & \multicolumn{2}{c}{Plot IV} & \multicolumn{2}{c}{Plot V} \\ \midrule
              & CD       & HD         & CD       & HD       & CD       & HD        & CD       & HD       & CD      & HD      \\ \midrule
Interpolation & 0.591    & 10.969     & 0.526    & 4.403    & 0.348    & 4.001     & 0.581    & 6.170    & 0.515   & 4.946   \\
PU-Net        & 0.452    & 7.699      & 0.294    & 3.441    & 0.427    & 4.703     & 0.343    & 6.864    & 0.453   & 5.056   \\
GradPU        & 0.577    & 10.970     & 0.514    & 4.426    & 0.337    & 4.039     & 0.542    & 6.170    & 0.485   & 4.950   \\
3DFSR (Ours)  & {\bf 0.410}   & {\bf 5.735}    & {\bf 0.254}   & {\bf 2.741}   & {\bf 0.249}   & {\bf 2.921}    & {\bf 0.328}   & {\bf 3.051}   & {\bf 0.263}  & {\bf 4.888}  \\ \bottomrule
\end{tabular}%
}
\end{table*}

Super resolution results using different methods for five plots are shown in Fig.~\ref{fig:sr_naive_plots} and Fig.~\ref{fig:sr_plots}. Overall, naive models can produce super-resolution point clouds, but have better results after using forestry data to train. The proposed 3DFSR results are visually closest to the reference data, exhibiting clean and uniform distributed point clouds with smooth ground and complete shapes of tree. Both bilinear interpolation and GradPU show artifacts and discontinuities of stems, and there are still noisy points, but perform well on ground points. PUNet produces point clouds with less discontinuities and relatively uniform point distribution, but they are noisy in both ground and trees compared to the other three methods.

Super resolution results at the single-tree level are shown in Fig.~\ref{fig:sr_singletree}. From Plot I to V, black walnut, oak, maple (\textit{Acer} spp.), fir (\textit{Abies} spp.), and pine are selected to represent the dominant species. 3DFSR can have complete tree shapes for all species. More importantly, only 3DFSR can produce complete circular shape of cross sections for stems, while other three barely recover cross sections and keep the original ALS point patterns.

We report quantitative results in Tab.~\ref{tab:sr_res} with different algorithms on five plots. The proposed 3DFSR consistently achieves the best performance in all plots for both Chamfer Distance (0.410, 0.254, 0.249, 0.328, and 0.263, respectively) and Hausdorff Distance (5.735, 2.741, 2.921, 3.051 and 4.888, respectively). bilinear interpolation obtains the highest Chamfer Distance and Hausdorff Distance in Plot I; 3DFSR obtains the lowest Chamfer Distance in Plot III and lowest Hausdorff Distance in Plot II. Among all four baselines, bilinear interpolation and GradPU obtain similar Chamfer Distance and Hausdorff Distance in all five plots, e.g., 0.592 and 0.577 Chamfer Distance in Plot I, and 4.001 and 4.039 Hausdorff Distance in Plot III. PU-Net generally gets lower Chamfer Distance and Hausdorff Distance than bilinear interpolation and GradPU in four plots except for Plot III, but still higher than 3DFSR.

\subsection{Stem detection}
\label{sec:stem_detect}
\subsubsection{Experimental setting}
In this experiment, we explore how super-resolution point clouds can improve the performance of stem detection. We run stem detection algorithms developed for ALS and TLS/MLS, respectively, on both original ALS point clouds and corresponding super-resolution point clouds. For ALS stem detection method, we use stem detection function in Digital Forestry Toolbox (DFT) developed by~\cite{parkan2018dft}. It follows the hypothesis that the point density of a stem position is higher than that in the surrounding area to detect stems. For TLS/MLS stem detection method, we use LeSSO developed by~\cite{shao2026three}. This is a stem detection algorithm by detecting circle shapes on the cross section at breast height. We conduct this experiment in the five plots mentioned in Section~\ref{sec:fi datasets}. For stem detection metrics, we use True Positive (TP), False Positive (FP), False Negative (FN), completeness, omission, commission and F1 score.

\subsubsection{Stem detection results}

\begin{table*}[t!]
\centering
\caption{Stem detection results using ALS-based method, DFT (Digital Forestry Toolbox), on both original ALS and SR ALS point clouds. ALS-based method can detect stems on both original ALS and SR point clouds, and obtain similar performance.}
\label{tab:dft_mapping}
\resizebox{\textwidth}{!}{%
\begin{tabular}{@{}lccccccc|ccccccc@{}}
\toprule
         & \multicolumn{7}{c|}{ALS-based method on ALS point clouds}        & \multicolumn{7}{c}{ALS-based method on super-resolution ALS point clouds}      \\ \midrule
         & TP & FP & FN & Completeness & Omission & Commission & F1   & TP & FP & FN & Completeness & Omission & Commission & F1   \\ \midrule
Plot I   & 91 & 54 & 0  & 1.00         & 0.00     & 0.37       & 0.77 & 91 & 60 & 0  & 1.00         & 0.00     & 0.40       & 0.75 \\
Plot II  & 20 & 9  & 10 & 0.67         & 0.33     & 0.31       & 0.68 & 23 & 12 & 7  & 0.76         & 0.23     & 0.34       & 0.71 \\
Plot III & 62 & 47 & 18 & 0.78         & 0.23     & 0.43       & 0.66 & 61 & 59 & 19 & 0.76         & 0.24     & 0.49       & 0.61 \\
Plot IV  & 9  & 4  & 0  & 1.00         & 0.00     & 0.31       & 0.82 & 9  & 3  & 0  & 1.00         & 0.00     & 0.25       & 0.86 \\
Plot V   & 10 & 13 & 1  & 0.91         & 0.09     & 0.57       & 0.59 & 11 & 9  & 0  & 1.00         & 0.00     & 0.45       & 0.71 \\ \midrule
Overall  & 192& 127& 29 & 0.87         & 0.13     & 0.40       & 0.71 & 195& 143& 26 & 0.88         & 0.12     & 0.42       & 0.70 \\
\bottomrule
\end{tabular}%
}
\end{table*}

\begin{table}[t!]
\centering
\caption{Stem detection results using MLS-based method, LeSSO, on SR ALS point clouds. MLS-based method cannot detect stems on original ALS point cloud, but can produce results on SR point clouds.}
\label{tab:lesso_mapping}
\resizebox{\columnwidth}{!}{%
\begin{tabular}{@{}lccccccc@{}}
\toprule
         & \multicolumn{7}{c}{TLS/MLS-based method on super-resolution ALS point clouds} \\ \cmidrule(l){2-8} 
         & TP     & FP    & FN    & Completeness    & Omission   & Commission   & F1     \\ \cmidrule(l){2-8} 
Plot I   & 91     & 0     & 0     & 1.00            & 0.00       & 0.00         & 1.00   \\
Plot II  & 26     & 0     & 4     & 0.84            & 0.16       & 0.00         & 0.91   \\
Plot III & 76     & 1     & 4     & 0.95            & 0.05       & 0.01         & 0.97   \\
Plot IV  & 9      & 3     & 0     & 1.00            & 0.00       & 0.25         & 0.86   \\
Plot V   & 10     & 1     & 1     & 0.91            & 0.09       & 0.09         & 0.91   \\
Overall  & 212    & 5     & 9     & 0.96            & 0.04       & 0.02         & 0.97   \\ \bottomrule
\end{tabular}%
}
\end{table}

We present stem mapping results obtained using the ALS-based DFT toolbox on both the original ALS point clouds and the super-resolution point clouds in Tab.~\ref{tab:dft_mapping}, and results of TLS/MLS-based method, LeSSO, in Tab.~\ref{tab:lesso_mapping}. Since LeSSO is designed for high-density point clouds, it is only applied to the super-resolution data.

For the results obtained with the DFT toolbox, the overall performance shows marginal changes when moving from the original ALS data to the super resolution point clouds, with F1 score from 0.71 to 0.70. In terms of completeness, Plot I and Plot IV both achieve 1.00 for both data types, Plot III decreases slightly from 0.78 to 0.76, while Plot II and Plot V show improvements (from 0.67 to 0.76 and from 0.91 to 1.00, respectively). On the original ALS point clouds, the method produces a large number of FP, leading to high commission errors. On the super-resolution point clouds with higher point density, it generates more FP in four out of five plots, resulting in further increased commission errors. In terms of F1 scores, Plot I and III decrease slightly, while Plot II, IV, and V show improvements.

LeSSO achieves higher performance (0.97 F1) than the DFT toolbox on the super-resolution point clouds (0.70 F1), but fails to produce results on the original ALS point clouds. Except for slightly lower completeness and higher omission in Plot V, LeSSO performs better (Plot I, II, and III) or similarly (Plot IV) than the DFT toolbox across all other metrics on the super-resolution point clouds. In addition to higher TP counts, the most notable difference is that LeSSO produces fewer FP than the DFT toolbox. As a result, LeSSO achieves completeness values ranging from 0.84 to 1.00 across all five plots, with low commission errors in most cases (0.00–0.09 in Plot I, II, III, and V, while Plot IV remains at 0.25). The corresponding F1 scores are higher than those of the DFT toolbox, reaching 1.00 in Plot I and exceeding 0.90 in Plot II, III, and V.

\subsection{DBH estimation}
\label{sec:dbh_estimate}
\subsubsection{Experimental setting}
In this experiment, we explore how super-resolution point clouds influence DBH estimation. For DBH estimation from ALS point clouds, previous studies commonly apply the implicit method, i.e. allometry equations. And for point clouds with a clear circle shape on the cross section, circle fitting is the common method. Therefore, for original ALS point clouds, we use allometry equation to estimate DBH; for corresponding super-resolution point clouds, we use circle fitting results to estimate DBH. This experiment is also conducted on the five plots mentioned in~\ref{sec:fi datasets}. To evaluate the accuracy of DBH estimation, we use bias, Mean Absolute Error (MAE), Root Mean Square Error (RMSE), and R\textsuperscript{2}.

\subsubsection{DBH results}
\begin{figure*}[]
	\centering
	\includegraphics[width=.99\textwidth]{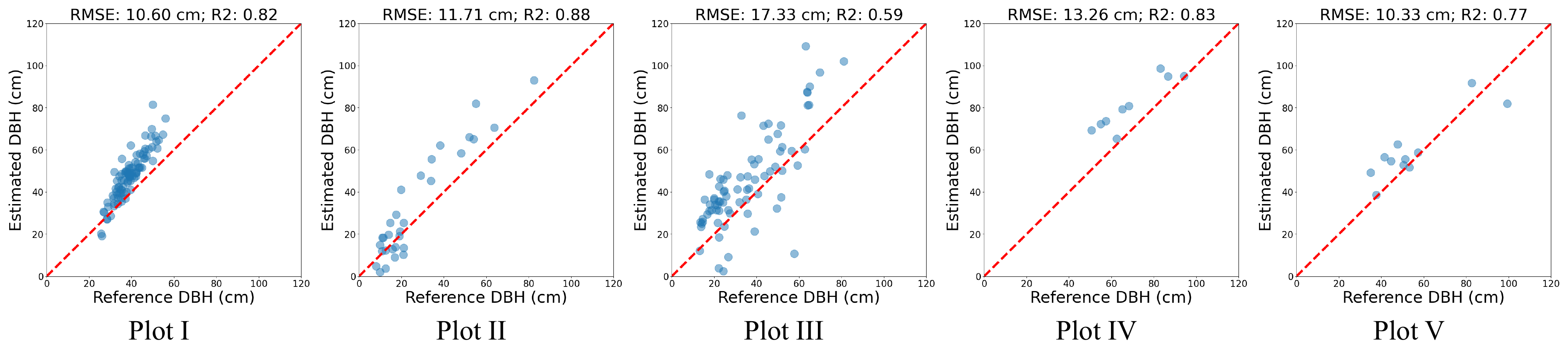}
	\caption{DBH estimation with original ALS point clouds using allometric equations.}
	\label{fig:dbh_allometry}
\end{figure*}
\begin{figure*}[]
	\centering
	\includegraphics[width=.99\textwidth]{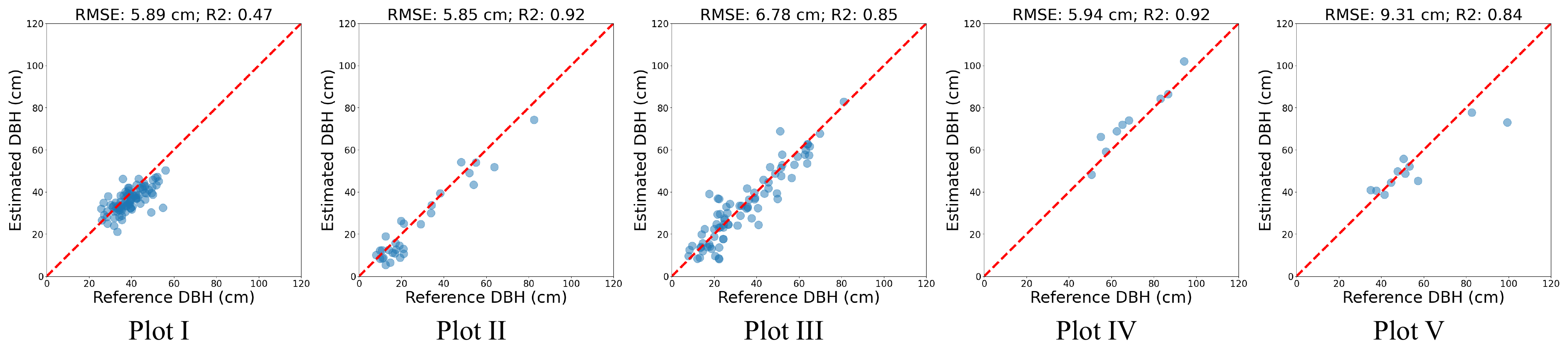}
	\caption{DBH estimation with super-resolution ALS point clouds using circle fitting.}
	\label{fig:dbh_sr}
\end{figure*}
\begin{table}[t!]
\centering
\caption{DBH estimation results on ALS point clouds with allometric equations.}
\label{tab:dbh_allometry}
\begin{tabular}{@{}lcccc@{}}
\toprule
         & Bias (cm)  & MAE (cm)   & RMSE (cm) & R\textsuperscript{2} \\ \midrule
Plot I   & 8.75  & 9.12  & 10.60 & 0.82  \\
Plot II  & 5.97  & 9.47  & 11.71 & 0.88  \\
Plot III & 9.70  & 14.38 & 17.33 & 0.59  \\
Plot IV  & 11.77 & 11.77 & 13.26 & 0.83  \\
Plot V   & 4.84  & 8.32  & 10.33 & 0.77  \\ \midrule
Overall  & 8.50  & 10.92 & 13.45 & 0.76  \\
\bottomrule
\end{tabular}%
\end{table}
\begin{table}[t!]
\centering
\caption{DBH estimation results on super-resolution ALS point clouds with circle fitting.}
\label{tab:dbh_sr}
\begin{tabular}{@{}lcccc@{}}
\toprule
        & Bias (cm)   & MAE (cm)  & RMSE (cm) & R\textsuperscript{2} \\ \midrule
Plot I  & -2.94  & 4.53  & 5.89  & 0.47   \\
Plot II & -2.86  & 4.87  & 5.85  & 0.92   \\
Plot III& -0.89  & 5.02  & 6.78  & 0.85   \\
Plot IV &  4.26  & 4.83  & 5.94  & 0.92   \\
Plot V  & -3.02  & 6.01  & 9.31  & 0.84   \\ \midrule
Overall & -1.90  & 4.84  & 6.43  & 0.87   \\
\bottomrule
\end{tabular}%
\end{table}

The relationship between the estimated DBH derived from the original ALS point clouds using allometric equations and the reference DBH in Fig.~\ref{fig:dbh_allometry}, and detailed quantitative metrics are reported in Tab.~\ref{tab:dbh_allometry}. Across the five plots, the allometry-based method shows biases between 4.84 and 11.77 cm, MAE ranging from 8.32 to 14.78 cm, and RMSE from 10.33 to 17.33 cm. Fig.~\ref{fig:dbh_allometry} exhibits a certain degree of overestimation, which is consistent with the positive bias values reported in Tab.~\ref{tab:dbh_allometry}. The R\textsuperscript{2} values range from 0.59 to 0.88, indicating moderate to strong correlations depending on the plot conditions.

Correspondingly, we present scatter plots of DBH obtained from super-resolution ALS point clouds using circle fitting, and show quantitative results in Tab.~\ref{tab:dbh_sr}. In general, the differences between the estimated DBH based on super-resolution ALS and the reference values are smaller than those obtained with the allometry-based method (Tab.~\ref{tab:dbh_allometry} and Tab.~\ref{tab:dbh_sr}). MAE is improved from 10.92 to 4.84 cm, and RMSE is changed from 13.45 to 6.43 cm. The biases range from -3.02 to 4.26 cm across the five plots, with only one plot showing a positive bias (4.26 cm). The MAE values range from 4.53 to 6.01 cm, and the RMSE values range from 5.85 to 9.31 cm, which is consistent with the fact that most samples lie close to the identity line in Fig.~\ref{fig:dbh_sr}. The R\textsuperscript{2} values reach 0.92 in Plot II and IV and remain above 0.84 in most plots, except for Plot I (0.47).

\subsection{Stem reconstruction}
\subsubsection{Experimental setting}
To further verify whether super-resolution point clouds can be used to obtain more detailed attributes, we reconstruct stems on both super-resolution point clouds and high-resolution MLS point clouds. If stems can be reconstructed from super-resolution point clouds and reconstructed stems have similar geometry with stems reconstructed from high-resolution point clouds, it means super-resolution point clouds can be used for stem analysis. For reconstruction method, we choose Sector Median Points (SMP) proposed by~\cite{shao2026three}. SMP has been proven to have the capability to reconstruct precise stem shapes on TLS/MLS point clouds. To see the difference between stems reconstructed from super-resolution point clouds and high-resolution point clouds, we first compare geometry quality using Chamfer Distance and Hausdorff Distance, and then show stem volume differences using MAE, RMSE and R\textsuperscript{2}. It should be noted that SMP was developed for temperate deciduous trees, so we select Plot I, II, III to conduct this experiment, as Plot IV and V are coniferous or mixed coniferous.
\subsubsection{Stem reconstruction results}
\begin{table}[t!]
\centering
\caption{Stem reconstruction performance in terms of Chamfer Distance (CD), Hausdorff Distance (HD), Mean Absolute Error (MAE), Root Mean Square Error (RMSE) and R\textsuperscript{2} computed by reconstructed stems from MLS point clouds and super-resolution point clouds.}
\label{tab:recon}
\resizebox{\columnwidth}{!}{%
\begin{tabular}{@{}lccccc@{}}
\toprule
         & CD (m)& HD (m) & MAE (m\textsuperscript{3}) & RMSE (m\textsuperscript{3}) & R\textsuperscript{2} \\ \midrule
Plot I   & 0.117 & 0.225 & 0.11 & 0.22 & 0.98 \\
Plot II  & 0.206 & 0.427 & 0.15 & 0.22 & 0.95 \\
Plot III & 0.220 & 0.535 & 0.12 & 0.18 & 0.80 \\ \midrule
Overall  & 0.170 & 0.377 & 0.13 & 0.20 & 0.96 \\
\bottomrule
\end{tabular}%
}
\end{table}
\begin{figure*}[]
	\centering
	\includegraphics[width=.99\textwidth]{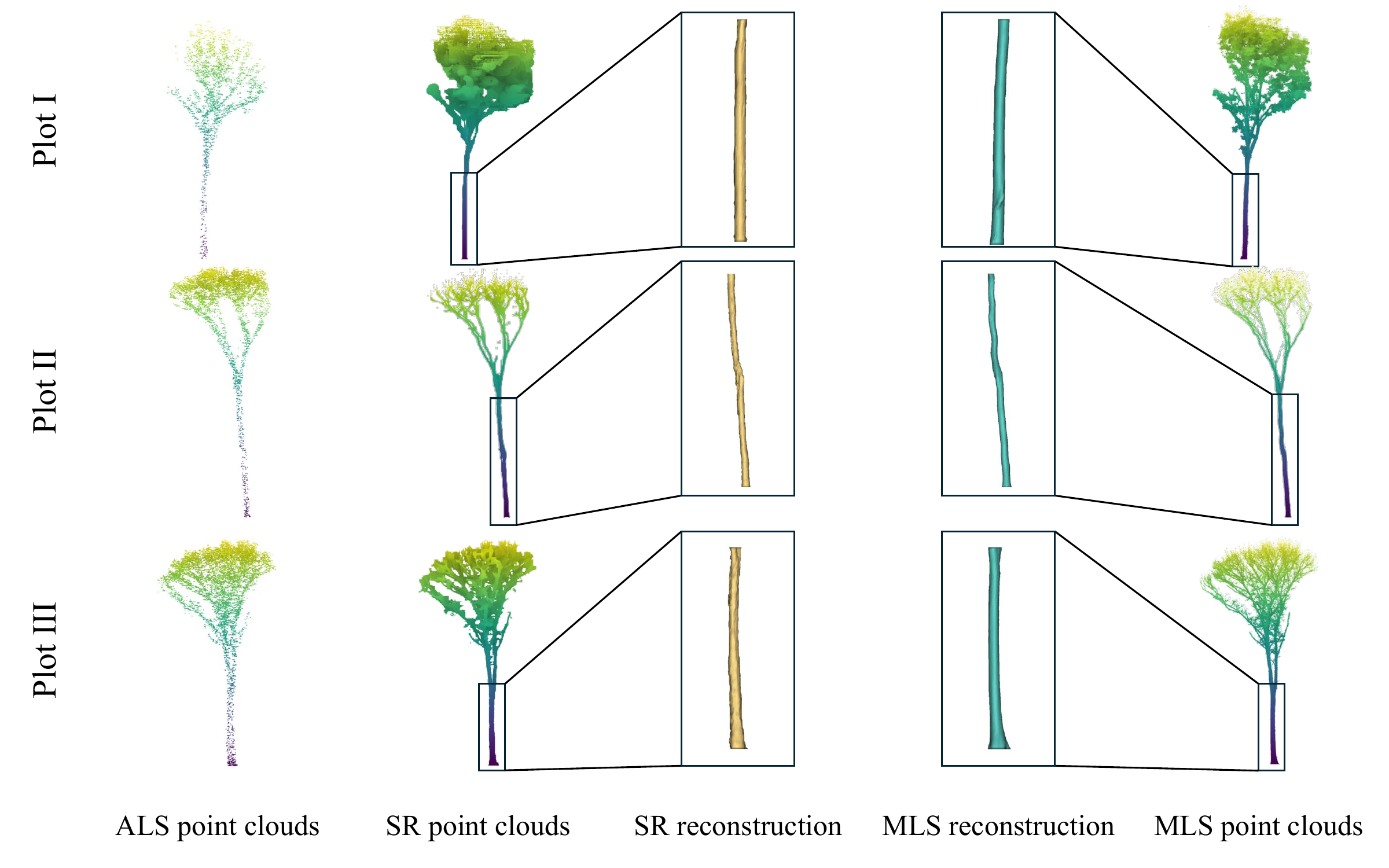}
	\caption{Stem reconstruction based on super-resolution (SR) point clouds and Mobile Laser Scanning (MLS) point clouds.}
	\label{fig:stem_recon}
\end{figure*}
\begin{figure*}[]
	\centering
	\includegraphics[width=.99\textwidth]{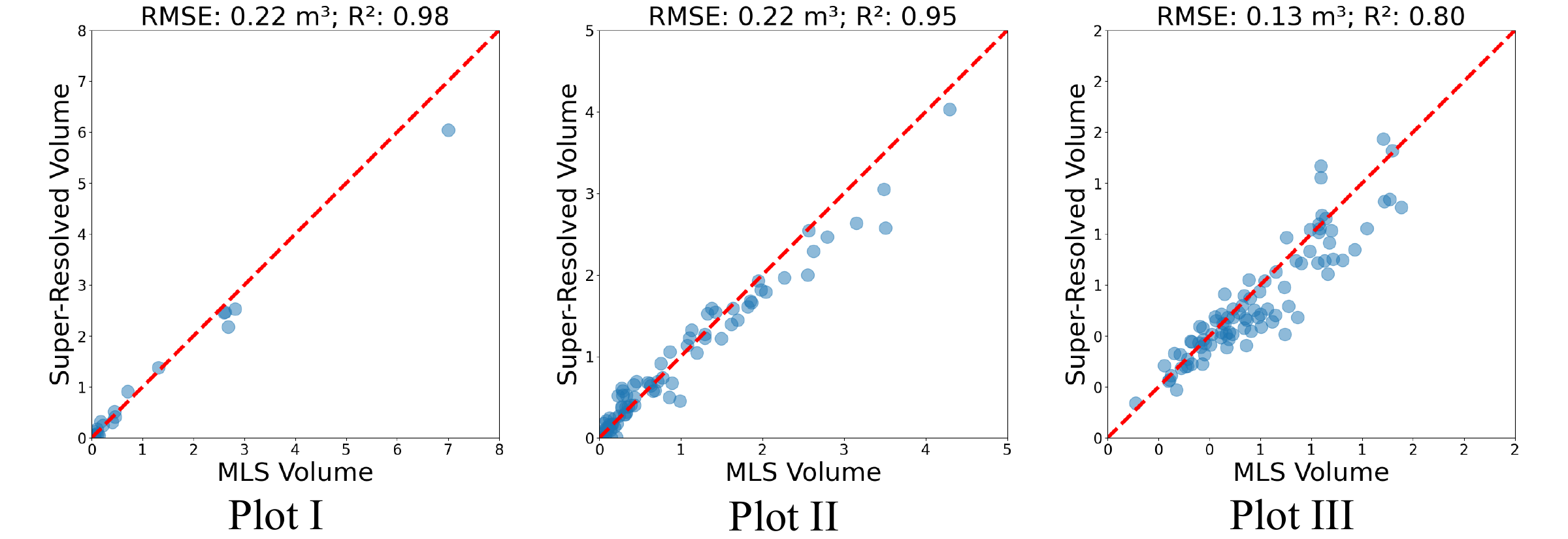}
	\caption{Stem volume comparison between stem reconstructed from super-resolution point clouds and MLS point clouds.}
	\label{fig:volume}
\end{figure*}

Stem reconstruction using MLS point clouds and super-resolution point clouds are shown in Fig.~\ref{fig:stem_recon}; and the qualitative results are reported in Tab.~\ref{tab:recon}. Experimental results show that the super-resolution point clouds can be successfully used to reconstruct stems. For the geometric evaluation, the Chamfer Distance and Hausdorff Distance indicate that reconstructed stems from super-resolution point clouds match those from MLS point clouds. Specifically, the overall Chamfer Distance and Hausdorff Distance are 0.170 m and 0.377 m, respectively. Among the three plots, Plot I achieves the best performance with 0.117 m of Chamfer Distance and 0.225 m of Hausdorff Distance, while Plot III shows the largest discrepancy with 0.220 m of Chamfer Distance and 0.535 m of Hausdorff Distance. In terms of stem volume estimation, we plot the relationship in Fig.~\ref{fig:volume}, stems reconstructed from super-resolution point clouds have consistent volume with stems reconstructed from high-resolution MLS point clouds. The overall MAE and RMSE are 0.13 m\textsuperscript{3} and 0.20 m\textsuperscript{3}, and R\textsuperscript{2} reaches 0.96, indicating stem geometry and volume can be extracted from super-resolution point clouds.

\subsection{Variation of input point densities}
\subsubsection{Experimental setting}
This experiment explores whether the proposed model can take data with different point densities as input, and how super-resolution results and their forest inventory results vary with different input point densities . The data for this experiment are from Plot VI. To ensure density is the only variable in this experiment, we chose to downsample the UAV LiDAR point clouds to different level of point densities and use them as input separately. Downsampled point densities are set to 10, 25, 50, 75, 100, 250, 500, 1000, 1500 and 1700. This range of point densities covers from the linear-model ALS data (10 points/m\textsuperscript{2}) to high-resolution UAV LiDAR data (1700 points/m\textsuperscript{2}). For all models in this experiment, we train models using one same georeferenced high-resolution MLS point cloud as reference and same hyperparameter settings mentioned in Section~\ref{method}. Stem detection and DBH estimation methods used in this experiment are the same as the methods in Section~\ref{sec:stem_detect} and Section~\ref{sec:dbh_estimate}. Chamfer Distance and Hausdorff Distance are used to evaluate the super-resolution performance, completeness, omission, commission and F1 are used to evaluate stem detection performance, and MAE and RMSE are used to evaluate DBH estimation performance.

\subsubsection{Density variation results}
\label{sec:density variation}
\begin{figure*}[]
	\centering
	\includegraphics[width=.99\textwidth]{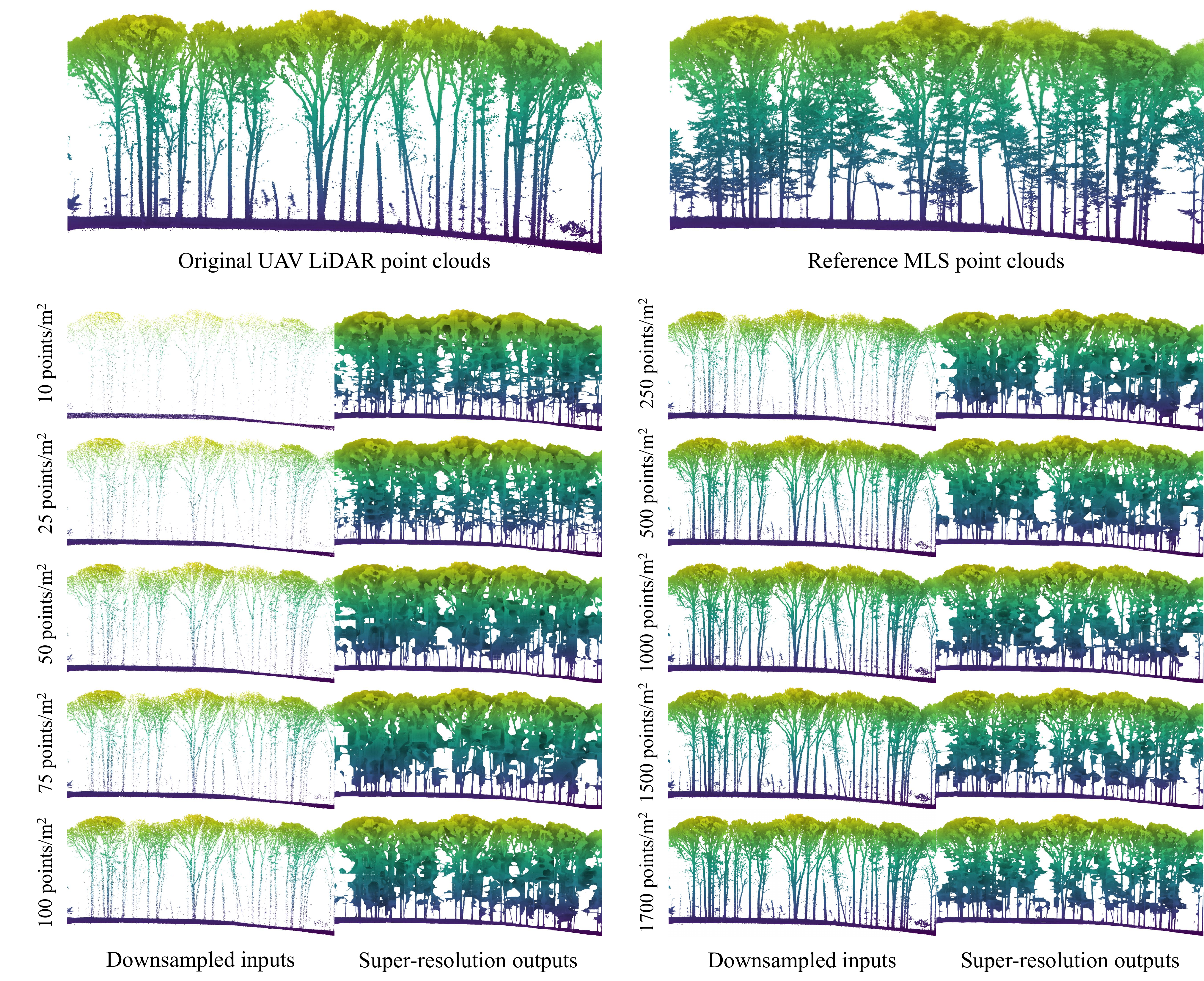}
	\caption{Super-resolution results using different densities as the input.}
	\label{fig:density_vis}
\end{figure*}
\begin{figure*}[]
	\centering
	\includegraphics[width=.99\textwidth]{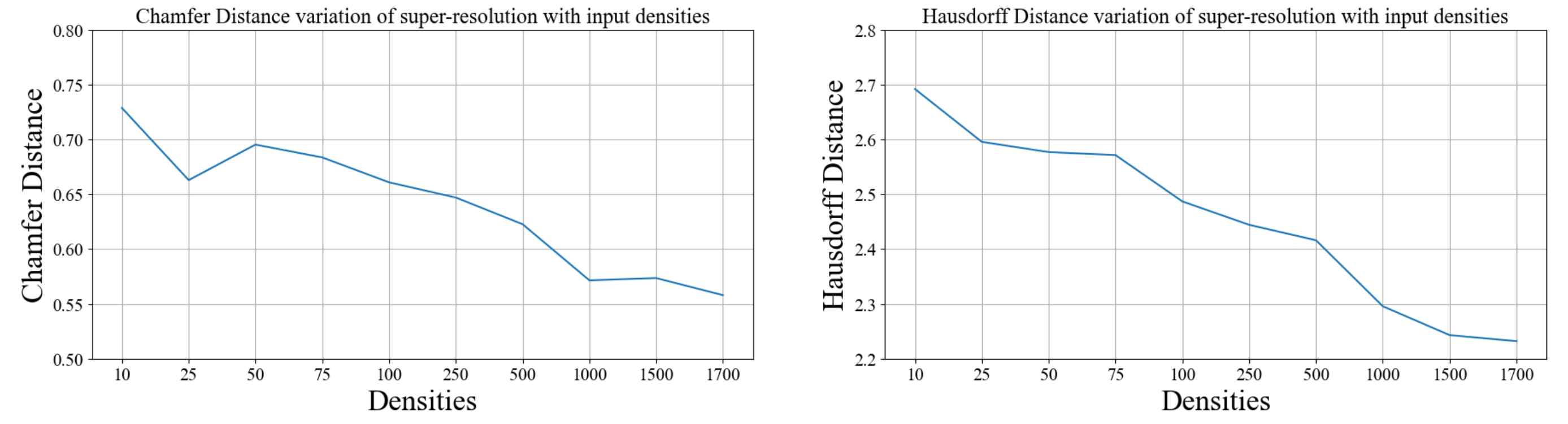}
	\caption{Performance variation on Chamfer Distance (left) and Hausdoff Distance (right) based on the change of input point densities.}
	\label{fig:density_trend}
\end{figure*}
\begin{figure*}[]
	\centering
	\includegraphics[width=.99\textwidth]{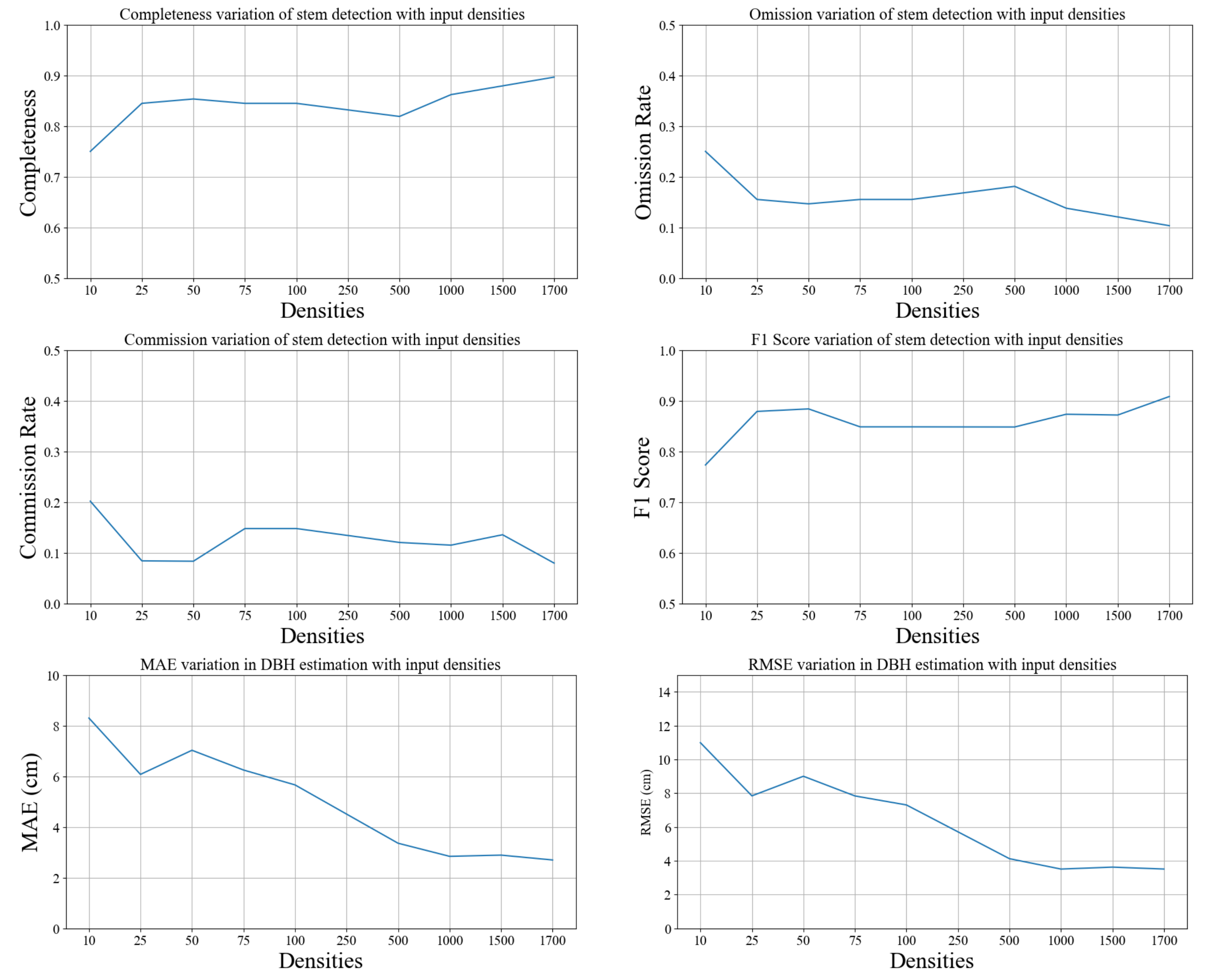}
	\caption{Performance variation on Chamfer Distance (left) and Hausdorff Distance (right) based on the change of input point densities.}
	\label{fig:density_tasks_trend}
\end{figure*}

\begin{table}[t!]
\centering
\caption{The point cloud super-resolution performance varying with changes of input point densities in terms of Chamfer Distance (CD), Hausdorff Distance (HD) and output point densities.}
\label{tab:density}
\begin{tabular}{lcc}
\toprule
\begin{tabular}[c]{@{}l@{}}Input densities \\ (points/m\textsuperscript{2})\end{tabular} & CD (m) & HD (m) \\ \midrule
10                                                                                                        & 0.729  & 2.692  \\
25                                                                                                        & 0.663  & 2.596  \\
50                                                                                                        & 0.695  & 2.577  \\
75                                                                                                        & 0.684  & 2.572  \\
100                                                                                                       & 0.660  & 2.487  \\
250                                                                                                       & 0.647  & 2.444  \\
500                                                                                                       & 0.622  & 2.416  \\
1000                                                                                                      & 0.571  & 2.296  \\
1500                                                                                                      & 0.573  & 2.243  \\
1700                                                                                                      & 0.558  & 2.232  \\ \bottomrule
\end{tabular}%
\end{table}

We show super-resolution results in Fig.~\ref{fig:density_vis}, and report qualitative results obtained different input point densities in Tab.~\ref{tab:density}, plot the super-resolution performance trends in Fig.~\ref{fig:density_trend} and plot stem detection and DBH estimation performance trends in Fig.~\ref{fig:density_tasks_trend}. Overall, the model can successfully produce super-resolution point clouds for all input densities. All generated super-resolution point clouds have dense and clean stem points, but have more clear canopy structures as the input densities increased. For some crowded areas, there are dense points between stems, and this also becomes better with point density increased. In terms of super-resolution quality, the model has the highest Chamfer Distance of 0.729 m and Hausdorff Distance of 2.692 m for 10 points/m\textsuperscript{2}, and lowest Chamfer Distance of 0.558 m and Hausdorff Distance of 2.232 m for 1700 points/m\textsuperscript{2}. With the increase of input point densities, as shown in Fig.~\ref{fig:density_trend}, the overall trends of Chamfer Distance and Hausdorff Distance gradually decrease, meaning the super-resolution qualities are gradually improved. The performance of stem detection and DBH estimation increases with higher input point densities. Specifically, for stem detection, super-resolution point clouds from 10 points/m\textsuperscript{2} have completeness, omission, commission, and F1 of 0.75, 0.25, 0.20, and 0.77, respectively; conversely, super-resolution point clouds from 1700 points/\textsuperscript{2} have 0.90, 0.10, 0.08, and 0.91 for those corresponding metrics. For DBH estimation, MAE decreases from 8.30 cm to 2.70 cm, while RMSE decreases from 11.01 cm to 3.51 cm.

\subsection{Model generalization}
\subsubsection{Experimental setting}
We further validate the model generalization capability across different LiDAR platforms. Specifically, we want to see whether a model trained on a LiDAR data obtained from one platform could be applied to unseen LiDAR data from another platform without transfer learning. To this end, we conduct two experiments. In the first one, we use the model trained based on ALS data described in Section ~\ref{sec:sr_quality}, and test it on the original UAV LiDAR data used in Section~\ref{sec:density variation}. In the second experiment, we use the model trained using original UAV LiDAR data in Section~\ref{sec:density variation}, and test it on the ALS LiDAR data used in Section~\ref{sec:sr_quality}. Chamfer Distance and Hausdorff distance are used to evaluate the performance.
\subsubsection{Model generalization results}
\begin{table}[t]
\centering
\caption{Generalization performance of the super-resolution model across LiDAR platforms.}
\label{tab:lidar_generalization}
\begin{tabular}{lcc}
\hline
\textbf{Cross-platform setting} & CD (m) & HD (m) \\
\hline
Train on UAV, test on ALS & 0.635 & 2.583 \\
Train on ALS, test on UAV & 0.726 & 2.961 \\
\hline
\end{tabular}
\end{table}
Model generalization performance across LiDAR platforms is shown in Tab~\ref{tab:lidar_generalization}. Overall, the results indicate that the proposed model maintains a reasonable level of performance when applied to unseen LiDAR data from a different platform, even without any transfer learning or fine-tuning. When trained on UAV LiDAR data and tested on ALS data, the model achieves a Chamfer Distance of 0.635 m and a Hausdorff Distance of 2.583 m. Conversely, when trained on ALS data and evaluated on UAV LiDAR data, the performance slightly degrades, with a Chamfer Distance of 0.726 m and a Hausdorff Distance of 2.961 m. This suggests that the model trained on UAV data generalizes relative better to ALS data than vice versa.
\section{Discussion}

Using ALS point clouds to conduct large-scale forest analysis has both ecological and economic significance. However, ALS super-resolution in forest environments remains largely unexplored due to its complex and unordered structures and lower quality of data. In this study, we introduced the concept of point cloud super-resolution to forest environments for the first time, aiming to enhance the quality of LiDAR-derived forest point clouds and thereby improve the accuracy of downstream forest inventory tasks. The effectiveness of our algorithm has been demonstrated by comparing it to other representative methods on five datasets. In addition, after point cloud super-resolution, performances on stem mapping and DBH estimation have been improved, and stem can be reconstructed. To our knowledge, this is the first work exploring super-resolution technique on ALS point clouds to improve forest inventory.

\subsection{Compare with baselines}
\label{discus1}

Point cloud super-resolution or upsampling has emerged in recent years as a key task in point cloud analysis, with numerous studies proposing different algorithms to enhance sparsely sampled 3D data~\citep{zhang2022point, kwon2023deep}. Most research focuses on increasing point density ~\citep{qiu2022pu, qian2021pu}, but does not solve noise in scans, which is very common in real-world laser scanning~\citep{mura2015estimating}. To examine the effectiveness of these upsampling approaches in complex forest scenes of ALS data, we compare three representative baselines to our proposed method: an unsupervised approach, bilinear interpolation; a feature-expansion–based model, PU-Net~\citep{yu2018pu}; an interpolation-refinement–based algorithm, GradPU~\citep{he2023grad}. Experimental results show that PU-Net tends to increase point density uniformly but introduces more noise, whereas bilinear interpolation and GradPU produce noticeable holes. We attribute this to their underlying mechanisms: bilinear interpolation considers only nearest points within local neighborhoods, PU-Net performs minor perturbations through feature expansion on original points, while GradPU interpolates based on the nearest eight neighboring points without accounting for local density variations, which is common in data captured in forest environment. Overall, our method substantially improves point cloud quality and achieves the best performance. It is also worth noting that the chosen baselines all generate a fixed upsampling ratio, requiring a predefined hyperparameter to determine the increase in point numbers, whereas our method predicts voxel occupancy according to a target resolution, determining whether each voxel is empty or occupied. Finally, all methods, including our 3DFSR, show limited improvement in super-resolving crown structures, which we attribute to the highly complex geometry of tree crowns and the fact that the training data of ground truth do not fully capture their true structural variability.

\subsection{Impacts on stem detection}
\label{discus2}
Stem detection is a basic but important task in forest LiDAR point cloud analysis. For high-density TLS/MLS point clouds, most tree detection methods (e.g., 3DFin, LeSSO) rely on the identification of stem points because these data contain clear, contiguous stem surfaces, and a single cylindrical stem typically suffices to indicate the presence of a tree~\citep{laino20243dfin, shao2026three}. In contrast, ALS point clouds are much sparser and stem points are often missing or barely visible, making stem-based approaches developed for TLS/MLS ineffective. Consequently, tree detection algorithms developed for ALS depend primarily on crown segmentation and peak-point localization~\citep{dalponte2016tree, amiri2018adaptive, li2012new}. The stem detection method of DFT represents an effort to leverage the limited visibility of stem points by exploiting local point-density differences~\citep{parkan2018dft}, but this strategy often introduces many false positives because the predefined rules tend to mistakenly classify non-stem structures as stems. Our approach addresses this limitation by applying point cloud super-resolution to ALS data, enhancing both density and clarity so that originally sparse, irregular stem points become well-defined cylindrical patterns. This improvement enables TLS/MLS-based stem detection algorithms to operate directly on super-resolved ALS point clouds, substantially increasing detection accuracy. This is because these methods are designed to identify cylindrical stems or circular cross-sections that become much clearer after super-resolution. Moreover, we observe that the application of ALS-based methods such as DFT to the super-resolved point clouds does not markedly affect their performance, as our algorithm uniformly increases point density across ground, stem, and crown regions, preserving the density-ratio cues on which DFT relies.

\subsection{Implicit and explicit DBH estimation}
\label{discus3}
Due to the sparsity of ALS point cloud data, it is usually impossible to directly measure DBH. This is because stems are poorly sampled by airborne LiDAR and trunk points are often very sparse in low-density scans~\citep{zhou2025extraction}. To solve this question, allometric equations have been developed based on tree height and crown diameter for trees in different regions, forest types, and functional groups~\citep{jucker2017allometric}. The advantage of this implicit approach is that DBH can be estimated using tree height and crown diameter even if there are no clear stem points in the point cloud~\citep{salas2010modelling}. In this research, we improve the quality of ALS point cloud data by developing a deep learning based super-resolution method, making it denser and cleaner, so DBH can be directly measured by circle fitting. It should be noted that although our method enables explicit measurements of ALS point clouds after super-resolution, it does not mean that implicit predictions based on allometric equations are meaningless. The use of allometric equations does not need high quality data of ALS point clouds. Taking our research as an example, even though deep learning-based super-resolution can increase point cloud density, it is ineffective for the reconstruction of understory vegetation and stems that are not scanned. However, estimation based on allometric equations does not require the stem to be visible. This means that even with low-density ALS point clouds where the stem points are not visible, relatively accurate DBH estimates can still be obtained if individual tree can be recognized~\citep{dalponte2018predicting, htoo2025development}.

\subsection{Feasibility of stem reconstruction}

The low density of ALS data often presents a challenge in reconstructing tree stems. Most studies on individual stem reconstruction rely on high-resolution UAV LiDAR or TLS/MLS data, and reconstructed geometry can be used for explicit estimation of detailed attributes, such as volume or biomass~\citep{pyorala2018assessing, brede2019non, shao2026three}. For ALS data, common practices utilize metrics derived from ALS data to develop regression models for the estimation of volume or biomass~\citep{hollaus2007airborne, lindberg2012comparison}. Our experiment on stem reconstruction further validates that super-resolution point clouds can support advanced structural analysis, stem reconstruction algorithms developed for TLS/MLS can be directly used on it without tuning. Reconstructed stem geometry and volume estimation results are consistent with MLS-derived results. This finding suggests that super-resolution of ALS point cloud technique could be used for individual stem level inventory across large forest areas.

\subsection{Applicability of varying input point density}

The applicability to point clouds of varying input densities is crucial for forest point cloud analysis, as it determines whether an algorithm can be used on data collected from different LiDAR platforms. To investigate whether our proposed 3DFSR algorithm remains effective when applied to different input point densities, we conducted the experiment that downsample high-resolution UAV LiDAR point clouds to different point densities and use them as input to train models and perform super-resolution. The experimental results demonstrate that our proposed voxel-based 3DFSR algorithm can successfully generate high-resolution point clouds for all input point clouds with densities ranging from 10 to 1,700 points/m\textsuperscript{2}. In terms of performance, our experiments reveal that the performance of point cloud super-resolution improves as input point density increases. We attribute this to the fact that higher-density inputs provide more complete and continuous point coverage, providing the model with richer information to facilitate the learning of the tree structures and overall forest scene distribution. Studies have also explored the impact of input point densities on the performance of forest analysis~\citep{xiang2024automated, wielgosz2024segmentanytree}. Experiments conducted in these studies indicated that voxel-based deep learning methods can successfully segment individual trees from point clouds with different point densities. And they also observed that the performance of forest inventory based on segmentation results increases as point density. Collectively, our study and previous research demonstrate that voxel-based deep neural networks are capable of processing various forest point cloud analysis tasks (e.g., super-resolution and individual tree segmentation); furthermore, these networks remain effective across a wide of point densities, and the performance exhibits a positive correlation with point cloud density. These findings are beneficial for the subsequent development of algorithms for point clouds from multiple LiDAR platforms.

\subsection{Generalization across LiDAR platforms}

Applying well-trained models to unseen data of LiDAR platforms is important to expand the generalization. Research has explored this question on segmentation tasks~\citep{krisanski2021sensor, lu2025towards}, and we study this question on super-resolution. Our cross-platform experiments that while the model is trained on a specific LiDAR platform, it maintains reasonable performance when applied to unseen data from a different platform. The performance gap between the two cross-platform settings can be attributed to differences in point density, noise characteristics, and scanning geometry between UAV and ALS LiDAR systems. UAV LiDAR data typically exhibit higher density and more detailed local structures, which may help the model learn richer geometric representations that transfer more effectively to ALS data. In contrast, models trained on ALS data may struggle to capture the finer details present in UAV data, leading to relatively higher errors. This further proves that voxel-based neural networks can handle different point densities LiDAR data for forest analysis.

\subsection{Limitations and future work}
\label{discus4}
There are several limitations in this study. First, our method does not perform well for tree canopy. It produces artifacts in areas with dense tree branches. The main reason is that TLS/MLS does not capture noise-free canopy points, so the model can learn limited information from this condition. Second, the model can only refine existing points and cannot generate data for areas that have not been scanned (e.g. unscanned understory or stems). This means that if the input data only have canopy structures, such as 3DEP, the model may not work. Third, the efficiency of its large-scale application is unknown. Areas of all plots used in this study are relatively small compared to stand or landscape-level coverage, so its real-world, large-scale application may take a long time. Take Plot I as an example, model inference of super-resolution using NVIDIA A5000 GPU (24 GB) takes 19 mins on 0.8 ha of forest. In the future, we will focus on improving the application of our method to large-scale forest ALS point clouds.
\section{Conclusion}

We introduced a deep learning model for ALS forest point cloud super-resolution. This model aims to fill the gap in the inaccurate estimation of forest attributes due to the low point density of forest ALS point clouds. It produces dense and clean point clouds from sparse and noisy ALS point clouds. In the evaluation of five plots from two different types of forests, our approach achieved the best performance compared to other representative upsampling algorithms. More importantly, we found that stem detection algorithms developed for TLS/MLS can be directly applied to super resolution point clouds, and produce accurate results; and subsequently, DBH can be estimated by circle fitting on super-resolution point clouds, outperforming allometry based methods. Our study shows that ALS point clouds can be enhanced through learning-based super-resolution, improving the accuracy of forest attribute estimation.
\section{Acknowledgment}

This work was partially supported by the National Institute of Food and Agriculture, United States [grant number 2023-68012-38992].

\clearpage
\bibliographystyle{cas-model2-names}

\bibliography{main}





\end{document}